%% file: main.tex
\documentclass{ecai}
\usepackage{times}
\usepackage{graphicx}
\usepackage{latexsym}

\usepackage[utf8]{inputenc}
\usepackage{amsmath,amssymb}
\usepackage{xspace}
\usepackage[ruled,linesnumbered]{algorithm2e}
  \SetCommentSty{commentstyle}
\usepackage[style=base]{caption}
\usepackage[labelformat=simple]{subcaption}
  
\usepackage{booktabs}
\usepackage{nicefrac}
\usepackage{tikz}
  \usetikzlibrary{positioning,shapes,shadows,arrows,decorations.pathreplacing,patterns}
\usepackage{pgfplots}
  \pgfplotsset{compat = 1.8}
\colorlet{ycolor}{green!40!blue}
\colorlet{xcolor}{red}
\usepackage{hyperref}

\newtheorem{example}{Example}
\newtheorem{proposition}{Proposition}
\newtheorem{assumption}{Assumption}

\definecolor{niceblue}{rgb}{0.0, 0.4, 0.6}
\definecolor{nicegreen}{rgb}{0.33, 0.42, 0.2}
\definecolor{niceorange}{rgb}{0.96, 0.52, 0.26}

\newcommand{\R}{\ensuremath{\mathbb{R}}\xspace}
\newcommand{\N}{\ensuremath{\mathbb{N}}\xspace}
\newcommand{\classes}{\ensuremath{\mathcal{Y}}\xspace}
\newcommand{\Oof}{\ensuremath{\mathcal{O}}\xspace}

\newcommand{\observe}{\ensuremath{\mathop{\mathbf{watch}}}}
\newcommand{\classify}{\ensuremath{\mathop{\mathbf{classify}}}}
\newcommand{\abstractfun}{\ensuremath{\mathop{\mathbf{abstract}}}}

\newcommand{\asgn}{\ensuremath{\gets}\xspace}
\newcommand{\vx}{\ensuremath{\vec{x}}\xspace}
\newcommand{\vv}{\ensuremath{\vec{v}}\xspace}

\newcommand{\vb}{\ensuremath{\vec{b}}\xspace}
\newcommand{\vzero}{\ensuremath{\vec{0}}\xspace}
\newcommand{\data}[2]{\ensuremath{\langle #1, #2 \rangle}\xspace}
\newcommand{\watchedOutputs}{\ensuremath{W}\xspace}
\newcommand{\clusters}{\ensuremath{\mathcal{C}}\xspace}
\newcommand{\cluster}{\ensuremath{C}\xspace}

\newcommand{\nfrac}[2]{\ensuremath{\nicefrac{#1}{#2}\xspace}}
\newcommand{\transpose}{^\mathsf{T}\xspace}
\newcommand{\cI}[1][\raisebox{0.3mm}]{{\color{niceorange}#1{\scalebox{0.5}{\ensuremath{\blacksquare}}}}\xspace}
\newcommand{\cII}{{\color{nicegreen}\ensuremath{\bullet}}\xspace}
\newcommand{\cNovel}{{\color{black}\ensuremath{\star}}\xspace}
\newcommand{\nnI}{NN$_1$\xspace}
\newcommand{\nnII}{NN$_2$\xspace}
\newcommand{\legendRaw}[2]{\includegraphics[clip,trim=64.5mm #1 90mm #2]{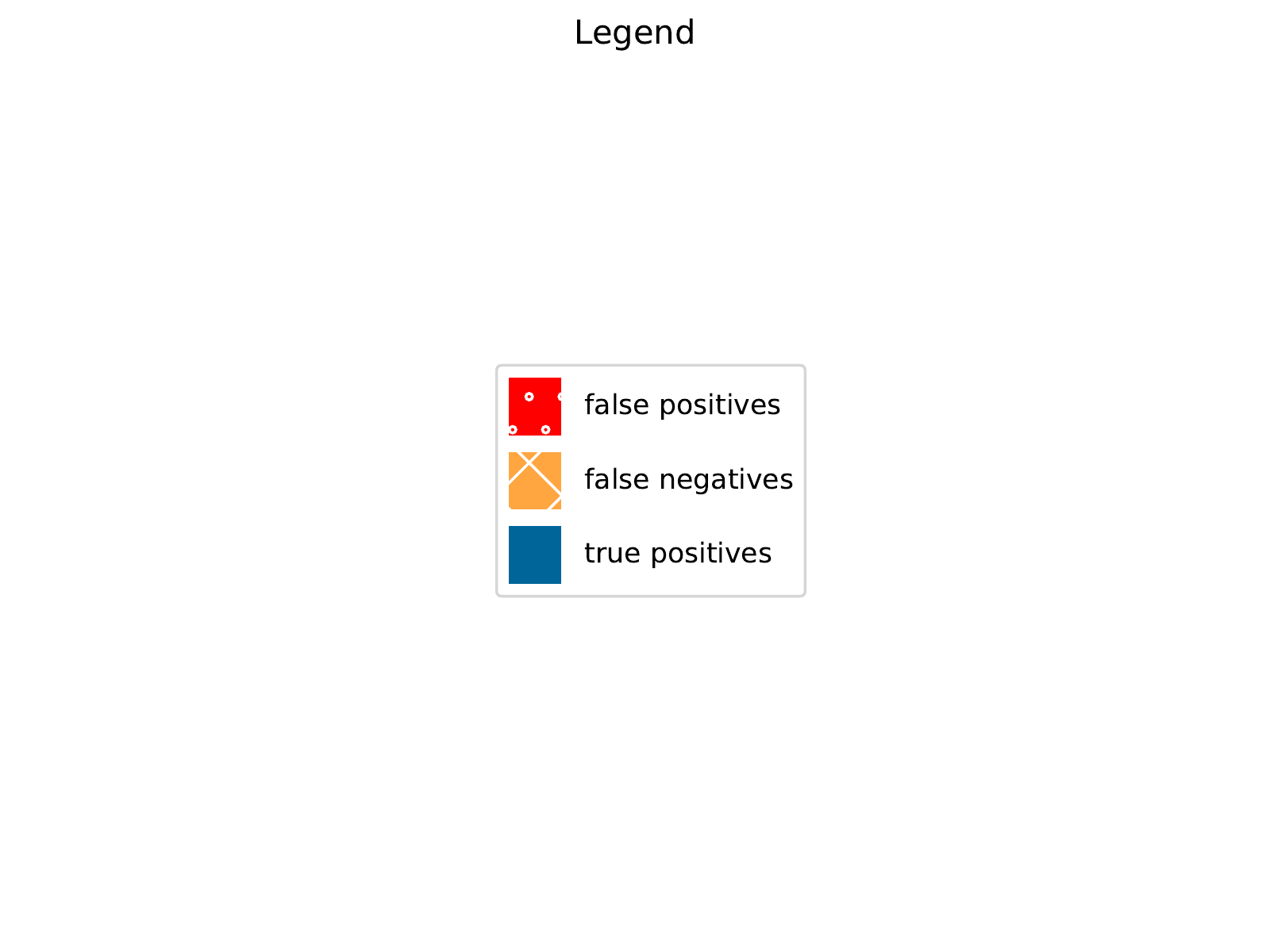}}
\newcommand{\legendTP}{\legendRaw{46mm}{67mm}}
\newcommand{\legendFN}{\legendRaw{56mm}{57mm}}
\newcommand{\legendFP}{\legendRaw{66mm}{47.5mm}}
\newcommand{\includecutplot}[3][13mm 8mm 21mm 18mm]{\includegraphics[clip,trim=#1,#2]{#3}}
\newcommand{\plotheight}{54.5mm}

\title{Outside the Box: \\ Abstraction-Based Monitoring of Neural Networks}

\author{Thomas A.\ Henzinger \and Anna Lukina \and Christian Schilling%
\institute{IST Austria, Austria, email: \{tah,anna.lukina,christian.schilling\}@ist.ac.at}}

\begin{document}

\maketitle

\begin{abstract}
\input{abstract}
\end{abstract}

\section{INTRODUCTION}

Neural networks have become the state of the art for a wide range of academic and industrial machine-learning applications, such as image or speech recognition~\cite{TouvronVDJ19,TanL19,YangDYCSL19}. With this technology becoming ever more widespread, one of the next great challenges is building techniques for identifying and mitigating intrinsic limitations of neural networks in the general problem domain of classification. Given an input, a neural-network classifier must, by definition, output one of the classes it was trained for. The ability to output ``\emph{do not know}'' for \emph{novel inputs} (i.e., inputs corresponding to classes the network was not trained for) is crucial for safety-critical applications. The software architects of autonomous cars, for instance, are facing a trade-off between efficiency and risk to misclassify anomalies~\cite{BojarskiTDFFGJM16,XuGYD17,Uber18}. This fundamental problem of novelty detection has been of great interest to the research community (see the survey~\cite{PimentelCCT14}). Moreover, evaluating learning algorithms in the face of parameter
or input variation has become a part of the emerging topic of explainable artificial intelligence~\cite{DoshiK17,MSKAY19}, where interpretability is investigated as a way to ascertain reliability of a learned system.

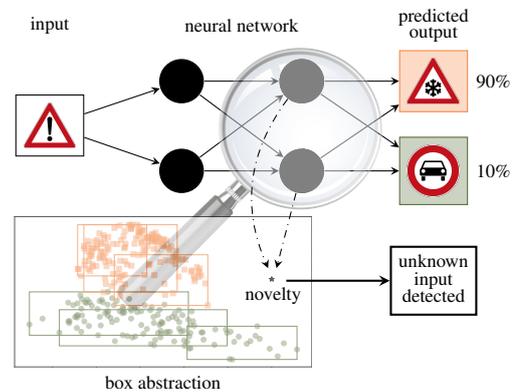
\begin{figure}[t]
    \centering
    \scalebox{0.98}{\input{tikz_framework}}
    \caption{A schematic view of our monitoring framework.}
    \label{fig:monitor_architecture}
\end{figure}

In search of a deeper understanding of the neural network's decision making and improved runtime management of the novel inputs, we turn to abstraction techniques commonly used in program analysis for monitoring complex safety-critical systems~\cite{CousotH78,Mine01,ClarkeHVB18,BartocciF18}.
Focusing on novelty detection, we propose to accompany the neural-network classifier with a \emph{runtime monitor} that supervises the decisions. Fig.~\ref{fig:monitor_architecture} depicts the high-level architecture of our framework. Running in parallel, the neural network and the monitor share the same interface, which allows for seamless integration into existing tools. The framework receives an input to be classified and can have two types of outputs: a classification (the neural network's decision) or a warning (``do not know'').

While this architecture is general enough to be applied to other classification techniques, our work is built around neural networks. The monitor ``watches'' a number of fixed network layers chosen to incorporate the essential feature information, namely layers close to the final network output~\cite{ZeilerF14}. The underlying assumption is that the neurons at the watched layers exhibit a pattern typical for inputs of the same class. The monitor is trained to recognize these patterns. At runtime, the output of the watched layers is compared against the corresponding pattern. In case of close resemblance, the monitor accepts the input and the framework outputs the class predicted by the neural network. In the opposite scenario, the monitor suspects the network of making a classification decision in an atypical way. With this suspicion, the monitor rejects the proposed class and outputs a warning about a possible novelty instead.

The patterns (abstractions) we consider are (unions of) intervals, or boxes, overapproximating the set of known neuron valuations. Despite their simplicity, our experiments show a remarkable novelty-detection performance. Owing to their efficiency, boxes can be used for runtime monitoring with no significant overhead.

\smallskip

\textbf{Our contributions} can be summarized as follows.
We propose an abstraction-based approach to detect novel inputs to neural-network classifiers, independent of their architecture.
The abstraction at chosen layers concisely represents all values ever seen during training.
We can efficiently identify novel inputs at runtime by comparing the behavior of the neural network to the abstraction.
Our approach can be tailored to a desired trade-off between the number of false warnings and undetected novelties.

\subsection{Related work}

\paragraph*{Runtime monitoring.}

Runtime monitoring for machine-learned systems is a common approach in the literature.
Bishop proposes to use a runtime monitor to estimate the uncertainty in a neural network, where statistical likelihood is used as the measure of novelty~\cite{Bishop93}.
However, unlike boxes, computing the likelihood is expensive.
The work by Gilpin considers a hierarchy of monitors: each component of a system has its own monitor, and for each subsystem consisting of several components there is a committee of monitors~\cite{Gilpin18}.
Dokhanchi et al.\ present \emph{quality temporal logic} for specifying properties of runtime monitors about label stability in video streams~\cite{DokhanchiADF18}.
Similar properties can be modeled with the \emph{model assertions} from~\cite{KangRBZ18}.
The abstraction in our approach could be used to explain when a label change is to be expected (namely, when, over time, the vectors observed at layer $\ell$ approach the border of the abstraction).
Cheng et al.\ introduce Boolean abstraction for neural-network monitoring, which, unlike the abstraction presented in this paper, is specific to ReLU activation functions (defined as $\sigma(x) = \max(x, 0)$ for neuron $x$)~\cite{ChengNY19}.
Due to the use of operations on Boolean logic with a binary decision diagram (BDD), their approach is only scalable for layers with a few neurons.

\paragraph*{Novelty detection.}

Novelty detection has been investigated by many researchers (see, e.g., \cite{PimentelCCT14} for a survey).
It is well known that the problem stems from differences in data distributions at training and prediction time~\cite{Ben-DavidBCKPV10,HendrycksG17}.
Some approaches, e.g., the work by Ganin and Lempitsky~\cite{GaninL15}, attempt to circumvent such cases by domain adaptation~\cite{PanY10}, which requires sampling the distribution at runtime.
Other approaches try to detect novelties probabilistically, e.g., using nonparametric density estimation~\cite{KnorrN97}.
Few approaches, e.g., \cite{RoyerL15}, perform an online adaptation of classifiers without having access to the whole distribution.
Our framework is orthogonal to these stochastic techniques, since we construct an abstraction.
Approaches such as \emph{$k$-centers}~\cite{YpmaD98} and \emph{support-vector data description}~\cite{NoumirHR12} consider ideas related to the ball abstraction presented in our experimental comparison (Section~\ref{sec:other_abstractions}), but they do not operate on neural networks.
Another solution to detect novelties is \emph{one-class classification}, where a classifier is trained to separate inputs of a single class from all other inputs (see, e.g., \cite{SabokrouKFA18} for a recent approach).

\paragraph*{Anomaly detection.}

In \emph{selective classification} (or \emph{abstention}) the idea is to reject inputs if a confidence score is too low~\cite{GeifmanE17,AlexandariSK18,JiangKGG18,ThulasidasanBBC19}.
Unlike novelty detection, selective classification is applied at training time already.
A well-known confidence score is the ``softmax score'' (see, e.g., \cite{HendrycksG17,GuoPSW17}), which we compare to in our experimental evaluation.
Liang et al.\ observe that novelties may result in lower confidence scores when applying \emph{supportive perturbation} to the input and \emph{temperature scaling} to the output~\cite{LiangLS18}. In contrast, our monitor does not require preprocessing of the input. Gal and Ghahramani illustrate limitations of the softmax output as confidence metric~\cite{GalG16}. Their approach requires access to the network structure to add dropout functions for modeling predictive uncertainty. Lakshminarayanan et al.\ quantify that uncertainty using ensembles of neural networks~\cite{Lakshminarayanan17}.

Sun and Lampert consider a generalization of novelty called \emph{out-of-specs situation}, which also includes the case that \emph{situations} from training never occur at runtime~\cite{SunL18}.
Although our approach targets the task of novelty detection only, according to the criteria proposed in that work, our framework is \emph{universal} (applicable to different network architectures), \emph{pre-trained ready} (requires no access to network training), and \emph{nonparametric} (uses no a priori knowledge about the data distribution), but not \emph{black-box ready} (since we require access to the network output in chosen layers).

Complementary to novelty detection, \emph{failure prediction} is the task of finding incorrect classifications that do \emph{not} arise from input novelty (see e.g., \cite{ZhangWFHP14}).
Some approaches, e.g., \emph{open-set learning}~\cite{BendaleB15} and \emph{zero-shot learning}~\cite{PalatucciPHM09} learn new classes at training respectively at prediction time, which also requires that novel inputs can be detected.

\section{PRELIMINARIES}
\label{sec:prelim}

We shortly introduce the basic terminology used in this paper.

\subsection{Neural networks}

\begin{figure}[t]
    \centering
    \begin{subfigure}{0.54\linewidth}
        \centering
        \hspace*{-3mm}
        \scalebox{0.8}{\input{tikz_neural_network}}
    \end{subfigure}
    \begin{subfigure}{0.45\linewidth}
        \centering
        \includecutplot[1mm 0mm 18mm 15mm]{keepaspectratio,width=\linewidth}{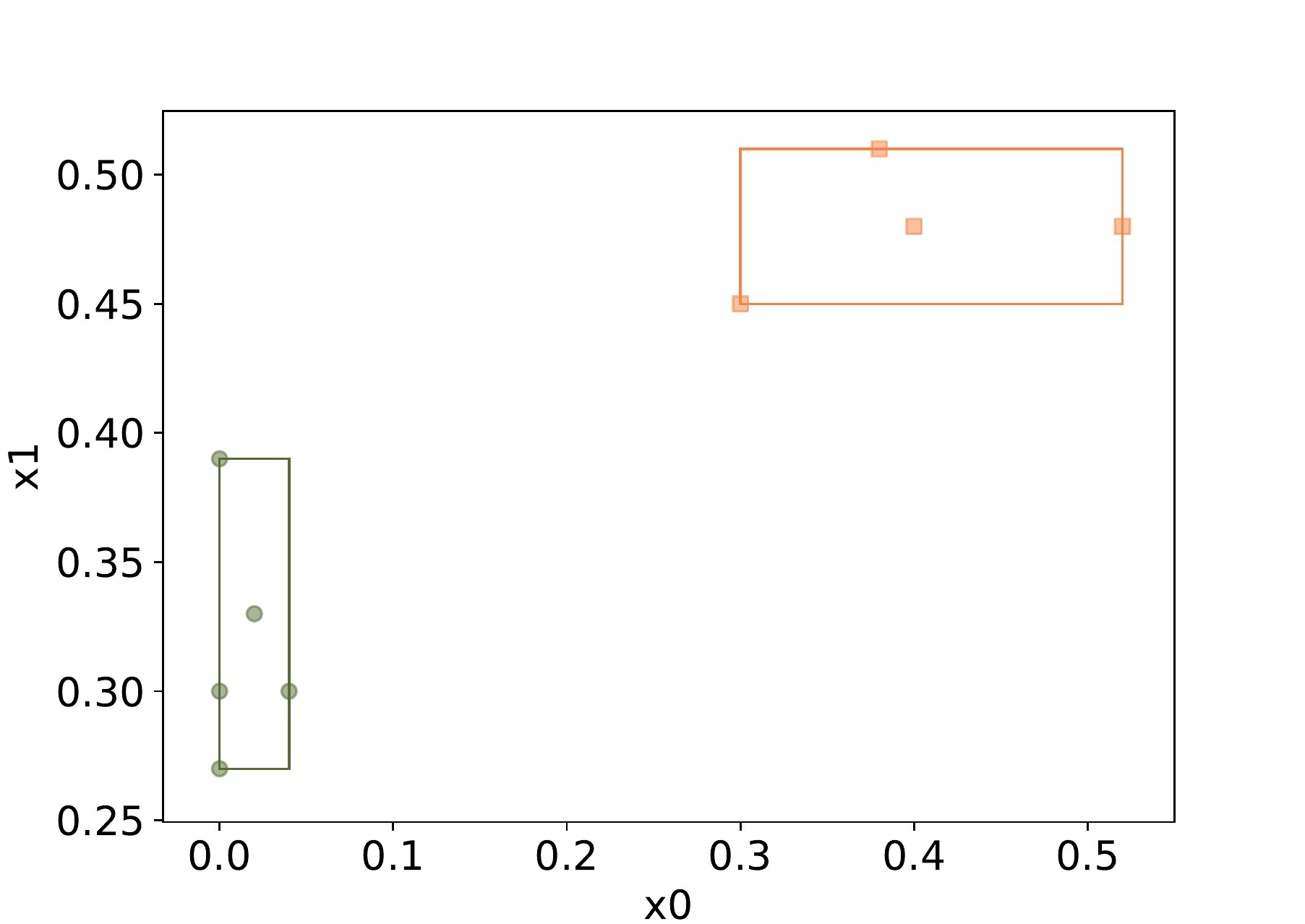}
    \end{subfigure}
    \caption{Left: example of a neural network that in each layer computes the function $f(\vx) = \sigma(A\vx+\vb)$ where $A$ given by the weights, $\vb = \vzero$, and $\sigma$ is the componentwise ReLU activation function.
    Right: box abstraction for layer $\ell_2$ and the inputs given in Ex.~\ref{ex:watching}.}
    \label{fig:nn_example}
\end{figure}

We assume the reader is familiar with the basic concepts of a neural network (see, e.g., \cite{Schmidhuber17}).
In Fig.~\ref{fig:nn_example} we depict a simple neural network, which we will later use to explain our approach.
We do not make any assumptions about the architecture or parameters (e.g., the activation functions) of the neural network.
Let \classes be the set of output classes.
Given an $n$-dimensional input $\vx \in \R^n$ and an index $\ell \in \N$ of a $d$-dimensional layer, we define the functions $\observe: \R^n \times \N \to \R^d$ and $\classify: \R^n \to \classes$ where $\observe(\vx, \ell)$ is the output at layer $\ell$ and $\classify(\vx)$ is the class predicted by the network.
Our approach is based on the following assumption, supported by other works~\cite{ZeilerF14}.

\begin{assumption}\label{asm:feature_layer}
    The layers close to the output layer of a neural network contain high-level information.
    Moreover, the output at these layers is similar for inputs of the same class and different for inputs of different classes.
\end{assumption}

\subsection{Box abstraction}

Our monitor will observe the output at a $d$-dimensional layer (i.e., it obtains vectors in $\R^d$).
Given a finite set $X \subseteq \R^d$ of such vectors, we want to construct a set $Y \supseteq X$ that generalizes $X$ to infinitely many elements.
This concept is known as \emph{abstraction}.
The rationale is to choose a simple representation for $Y$ that is easy to manipulate and answer queries for.
In this context, we are interested in the following operations on such sets $Y$:
\begin{itemize}
    \item creation of a set $Y$ from a finite set $X$ of vectors,

    \item a membership test for a vector \vx (i.e., deciding $\vx \in Y$), and

    \item (optional) enlargement (or bloating) to a superset $Y' \supseteq Y$.
\end{itemize}
We focus on the \emph{interval abstraction}~\cite{CousotC76} where the set $Y$ is a Cartesian product of intervals $[l_i, u_i]$ with $l_i$ and $u_i$ being the respective lower and upper bounds in dimension~$i$.
Geometrically, the shape of $Y$ is called a \emph{box} (or hyperrectangle).
A $d$-dimensional box can be represented by $2d$ bounds.
Creating a tight box around a set of $m$ vectors is a simple $\Oof(dm)$ task.
Testing membership of a vector in a box is in $\Oof(d)$.
Boxes can be enlarged (absolutely or relatively) by extending the bounds in $\Oof(d)$.
In this work, we propose an extension to a union of such boxes, which we call the \emph{box abstraction}.

\section{OUTSIDE-THE-BOX MONITORING}

In this section we describe the process of building and employing a monitor for a neural-network classifier.

\subsection{Constructing a monitor}

The first step is to construct a monitor for a given trained neural network and a labeled training dataset.
We would typically use the same dataset that the network was trained on, but this is not obligatory.
We require access to the network's output at predefined layers (which we call the \emph{watched layers}).
To simplify the presentation, we describe the concept for a single watched layer $\ell$, but we discuss the generalization to multiple layers in Section~\ref{sec:multiple_layers}.

\begin{algorithm}[tb]
    \caption{Constructing abstraction at layer $\ell$}
    \label{algo:train}
    \KwIn{%
    \classes: output classes \\
    $D = \{\data{\vx^{(1)}}{y^{(1)}}, \dots, \data{\vx^{(m)}}{y^{(m)}}\}$: training data \\
    $\tau$: clustering parameter
    }
    \vspace*{1mm}
    \KwOut{$A_1, \dots, A_{|\classes|}$: lists of abstractions}
    \vspace*{2mm}
    \For{$y \in \classes$}{
        \tcp{collect all outputs at layer $\ell$ for inputs of class $y$}
        $\watchedOutputs_y \asgn \left\{\!\observe(\vx, \ell)  \bigg| \data{\vx}{y} \in D \land y = \classify(\vx)\!\right\}$\label{line:collect}\!\;
        $\clusters_y \asgn$ cluster($\watchedOutputs_y$, $\tau$)\label{line:cluster}\tcp*[l]{divide collected vectors into clusters}
        $A_y \asgn [\;]$ \tcp*[l]{list of abstractions for class $y$}
        \For{$\cluster \in \clusters_y$}{
            \tcp{construct abstraction for vectors in cluster $\cluster$}
            $A_y^\cluster \asgn$ \abstractfun($\cluster$)\label{line:abstract}\;
            $A_y$.add($A_y^\cluster$) \tcp*[l]{add abstraction to list}
        }
    }
    \Return{$A_1, \dots, A_{|\classes|}$}
\end{algorithm}

In Algorithm~\ref{algo:train} we present the pseudocode for constructing a monitor at layer $\ell$.
The algorithm consists of three phases for each output class.
In the first phase (line~\ref{line:collect}), we run the network on the training data while watching layer $\ell$, i.e., we collect the corresponding output at layer $\ell$.
The output is labeled with the corresponding ground-truth class (we only consider correctly classified input data\footnote{For the purpose of novelty detection, whether or not to consider misclassified inputs is not crucial. However, experiments suggested that ignoring misclassified inputs improves the false-negative rate of the monitor.}).

\begin{example}\label{ex:watching}
    Recall the network from Fig.~\ref{fig:nn_example} and the following labeled training data for output classes \cI and \cII:
    \begin{align*}
        D = \{ \ \ & \data{(0.5, 0.5)\transpose}{\cI}, \data{(0.5, 0.6)\transpose}{\cI}, \data{(0.4, 0.6)\transpose}{\cI}, \\
        & \data{(0.2, 0.7)\transpose}{\cI}, \data{(0.7, 0.2)\transpose}{\cII}, \data{(0.6, 0.2)\transpose}{\cII}, \\
        & \data{(0.7, 0.1)\transpose}{\cII}, \data{(0.8, 0.1)\transpose}{\cII}, \data{(0.9, 0.2)\transpose}{\cII} \ \ \}
    \end{align*}
    Watching the second (hidden) layer $\ell_2$, we obtain the following vectors, which we label with the ground truth (also depicted in Fig.~\ref{fig:nn_example}):
    \begin{align*}
        \watchedOutputs_{\cI} &= \{ (0.3, 0.45)\transpose\!, (0.38, 0.51)\transpose\!, (0.4, 0.48)\transpose\!, (0.52, 0.48)\transpose \} \\
        \watchedOutputs_{\cII} &= \{ (0.02, 0.33)\transpose\!, (0.04, 0.3)\transpose\!, (0, 0.27)\transpose\!, (0, 0.3)\transpose\!, (0, 0.39)\transpose \}
    \end{align*}
\end{example}

Having obtained the (labeled) vectors from the training data, we continue with the second phase.
Since the vectors collected for each class often cover different regions of the state space, we use a clustering algorithm to group the vectors based on their region (line~\ref{line:cluster}).
We note that this step is not mandatory, but we found that it can improve the precision substantially.
In our implementation, we use \emph{$k$-means} clustering~\cite{Lloyd82}, which requires to fix the number of clusters in advance; hence we iteratively increase the number of clusters until the relative improvement of the sum-of-squares metric falls below a threshold $\tau$, which is an input parameter of the algorithm.

\smallskip

In the third phase, we construct a box abstraction for each combination of class and cluster identified before (function \abstractfun\ in line~\ref{line:abstract}).
As a result, we obtain a list of abstractions for each class.

\begin{example}\label{ex:abstractions}
    Consider again Example~\ref{ex:watching}.
    For simplicity, we assume that we obtain a single cluster for each class.
    For each class-cluster combination, we construct the tightest box that contains all vectors, as depicted in Fig.~\ref{fig:nn_example} on the right.
    For instance, for class~\cII the extremal values are $[0, 0.04]$ in dimension~0 and $[0.27, 0.39]$ in dimension~1, which correspond to the green box in the lower left.
\end{example}

\subsection{Monitoring procedure}

We have computed an abstraction by watching a given layer $\ell$ during monitor training.
We now describe how the monitor uses this abstraction to operate at runtime.

\begin{algorithm}[tb]
    \caption{Monitoring at layer $\ell$}
    \label{algo:run}
    \KwIn{%
    $\vx$: network input \\
    $A_1, \dots, A_{|\classes|}$: lists of abstractions
    }
    \vspace*{1mm}
    \KwOut{\textit{``accept''}/\textit{``reject''}: answer}
    \vspace*{2mm}
    $y \asgn$ \classify(\vx)\label{line:classify} \tcp*[l]{predict class of \vx}
    $\vv \asgn$ \observe(\vx, $\ell$)\label{line:watch} \tcp*[l]{collect output at layer $\ell$}
    \For(\tcp*[h]{check each abstraction for class $y$}){$A_y^C \in A_y$}{
        \If{$\vv \in A_y^C$\label{line:membership}}{
            \Return{``accept''}\label{line:accept} \tcp*[l]{found an abstraction containing \vv}
        }
    }
    \Return{``reject''}\label{line:reject} \tcp*[l]{\vv is not contained in any abstraction}
\end{algorithm}

Recall the architecture from Fig.~\ref{fig:monitor_architecture}.
We summarize the pseudocode of the monitoring procedure in Algorithm~\ref{algo:run}.
Given an input vector \vx, we first ask the network for a prediction $y$ (function \classify\ in line~\ref{line:classify}).
While the inputs are propagated through the network, we observe the output vector \vv at layer $\ell$ (line~\ref{line:watch}).
(Note that, in practice, the second step can be implemented as part of the first step instead of querying the neural network twice.)
We then ask the abstractions $A_y$ constructed for class $y$ whether one of them contains \vv (line~\ref{line:membership}).
If this is the case, we conclude that the network processed the input in a usual way, and the monitor accepts the prediction $y$ (line~\ref{line:accept}).
Otherwise, we conclude that the network has processed the input in an unusual way, and the monitor rejects the input (line~\ref{line:reject}).

\begin{figure}[t]
	\centering
	\includecutplot[3mm 0mm 20mm 18mm]{keepaspectratio,width=\linewidth,height=49mm}{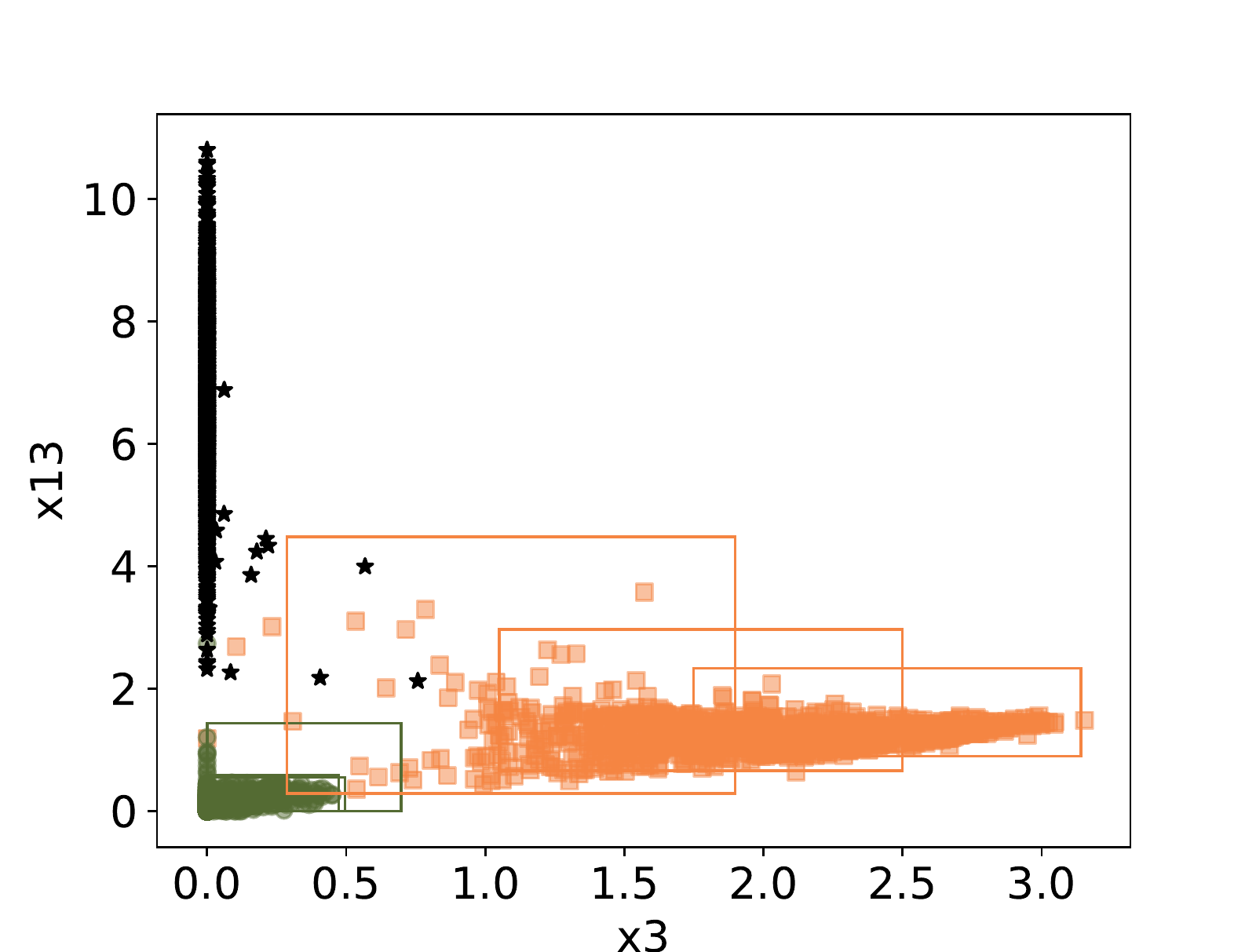}
	\caption{Box abstraction and the outputs obtained for new data at the second-to-last layer on the MNIST benchmark (see Section~\ref{sec:evaluation}) with two known classes (\cI, \cII) and one unknown class (\cNovel) (projection to two arbitrary dimensions).
	For better visibility, we fixed the number of boxes to three per class.}
	\label{fig:boxes}
\end{figure}

\begin{example}
	In Fig.~\ref{fig:boxes}, we give an example of the box abstraction in an arbitrary 2D projection for one of our benchmarks.
	Note that the boxes are created from the training dataset and the figure shows the points from the test dataset.
	With few exceptions, the boxes still contain the points of the same class.
	As can be seen in the figure, most of the novelties are not contained in any box for this projection, which leads to rejection by the monitor regardless of the class the network predicts.
	Assuming that the three novelties inside the projected box of class~\cI are also contained in the high-dimensional box, these novelties will still be rejected if the network predicts the class~\cII.
\end{example}

\section{DISCUSSION}

We now discuss the detection effectiveness, the extension to multiple layers, and the resource efficiency of our monitoring approach.

\subsection{Detection effectiveness}

We consider the potential outcomes of a query to a monitored network.
Firstly, since the abstractions \emph{over}approximate all vectors ever seen during training, the monitor never rejects a training input.

\begin{proposition}
    Given a monitored network and an input \vx used for monitor training, the monitor always accepts the prediction of the network for input \vx at runtime.
\end{proposition}

Additional training always raises the monitor's acceptance rate since new inputs can only increase the abstraction.
We say that an abstraction has \emph{converged} if it has reached a fixpoint, i.e., if additional training on any new input of a known class does not enlarge the abstraction anymore.
Monitors with converged abstraction generalize the above property to any inputs of the known classes, and conversely, warnings of monitors with converged abstraction are always genuine.
Consequently, false warnings can only occur for non-converged abstraction, and they indicate that more training data is required.
We summarize this observation below.

\begin{proposition}
    Consider a monitored network with converged abstraction and an input \vx.
    If the network predicts the correct class, the monitor accepts the prediction.
    If the monitor rejects the prediction, then \vx does not belong to the predicted class.
\end{proposition}

By design, an abstraction loses the direct link to the inputs it was created for. As a result, if a point is outside the abstraction, then we can determine with certainty that this point does not belong to the set outlined by the abstraction. On the other hand, when a point lies inside the boundaries of the abstraction, we cannot conclude if the set of values inside the abstraction and the given point belong to the same input class.
This property makes abstractions efficient, and determining the values outside an abstraction is sufficient for novelty detection.
However, if a novelty produces layer outputs falling inside an abstraction, we would still like to reject this input.
Here one could use another novelty-detection approach in parallel.

Assume the abstraction has not yet converged.
We differentiate two cases of false warnings (false positives).
In the first case, the vector obtained from the watched layer is outside the abstraction but close to its border.
Then the abstraction is too precise, meaning that its generalization power is too weak.
Generally one can enlarge the abstraction to prevent such scenarios.
In the second case, the vector is far outside the abstraction.
According to Assumption~\ref{asm:feature_layer} that the output at the layer $\ell$ is representative of the input, it follows that the network has processed the input in a unique way never observed during training.
Hence this case indicates insufficient training.

Now consider the case when the monitor accepts the input, i.e., the vector observed at layer $\ell$ is inside the abstraction, but the prediction of the network is wrong (false negative).
With Assumption~\ref{asm:feature_layer}, the monitor should not have seen any neighboring vector during training.
In this case, the abstraction is too coarse.
We proposed to use clustering in order to mitigate such scenarios.
As an alternative, we also experimented with more precise types of abstractions, but the results were not convincing (see Section~\ref{sec:other_abstractions}).

We note that for each class we monitor every neuron in the layer.
However, typically only a subset of the neurons (identifiable using principal component analysis~\cite{Jolliffe86}) is relevant for a given class.
Focusing on these neurons is beyond the scope of this work.

\subsection{Monitoring multiple layers}\label{sec:multiple_layers}

There are two main options to generalize the abstraction from one layer to multiple layers.
The first option is to consider a list of monitors, one for each layer, and to accept an input if and only if all monitors accept it.
The second option is to instead concatenate the outputs of the layers, i.e., for two layers $\ell_i$ and $\ell_j$ to consider $\observe(\vx, \ell_i) \cdot \observe(\vx, \ell_j)$ as the input to the (single) monitor.
In this work we use the first option, which we justify as follows.
While the second option is more precise as it keeps track of dependencies between neurons in different layers, the hierarchical structure of neural networks makes inter-layer dependencies less important.
The second option also increases the input dimension for the monitor, which leads to the curse of dimensionality, especially for clustering.

\subsection{Resource efficiency}

As mentioned in Section~\ref{sec:prelim}, the box abstraction provides linear-time operations for creation, membership, and (relative or absolute) enlargement of each box.
In particular, creating a box abstraction from samples is an incremental process, which is advantageous for processing large datasets, as we do not require to store all data at once.
Note that our presentation of Algorithm~\ref{algo:train} processes the data in one batch for clustering, but any online-clustering algorithm (e.g.,~\cite{CohenAddadGKR19}) can be used instead.
We remark that working with a box abstraction can also be parallelized, but we use a sequential implementation.

The size of memory required to store a box is independent of the amount of training data.
Thus the memory requirement is linear in the number of neurons of the watched layers and the number of clusters.

\section{EXPERIMENTAL EVALUATION}\label{sec:evaluation}

In this section, we assess the performance of our monitoring framework, where we employ the following setup.
To emulate the scenario of novel inputs, we train a neural network on $k$ out of $n$ available classes and vary $k$ in the range $[2, n)$.
We choose the first $k$ classes in the order defined by each dataset.
(Further experiments did not reveal a significant correlation between the monitoring performance and the particular ordering of the $k$ classes.)
We use the same network training parameters (e.g., number of epochs) for all instances of the same benchmark.
After training the network, we construct the monitor from the same training data.
To simulate convergence of the abstraction, we also include a scenario where we additionally train the box abstraction on the $k$ classes of the test dataset.
We then run the monitored network on the whole test dataset of $n$ classes.

Since our approach is complementary to classification techniques, we analyze four popular datasets for image classification: MNIST~\cite{lecun1998gradient}, fashion MNIST (F\_MNIST)~\cite{fashionMNIST}, CIFAR-10~\cite{Krizhevsky09}, and German traffic signs (GTSRB)~\cite{Stallkamp-IJCNN-2011}.
We use two neural-network models from~\cite{ChengNY19}, which we call \nnI and \nnII.
More information on the datasets and training parameters are provided in Tab.~\ref{tbl:network_accuracy}.

\begin{table}
    \caption{Benchmarks and their parameters used in training the models from~\cite{ChengNY19}.
    Column ``Acc.\ train / test'' shows the training and testing accuracy, respectively. Accuracy fluctuates for the models trained on different numbers of classes and we provide a range.}
    \label{tbl:network_accuracy}
    \centering
    \scriptsize{
    \def\d{@{\hspace*{1.75mm}}}
    \def\de{@{\hspace*{1mm}}}
    \begin{tabular}{\de c \d l \d c \d c \d c \d c \d l \de}
        \toprule
        ID & Dataset & Classes & Inputs train / test & Network & Epochs & Acc.\ train / test $\%$ \\
        \midrule
        1 & MNIST & 10 & 60,000 / 10,000 & \nnI & 10 & $99/99$ \\
        2 & F\_MNIST & 10 & 60,000 / 10,000 & \nnI & 30 & $98-99/91-99$ \\
        3 & CIFAR-10 & 10 & 60,000 / 10,000 & \nnII & 200 & $99/71-95$ \\
        4 & GTSRB & 43 & 39,209 / 12,630 & \nnII & 10 & $98-99/88-97$ \\
        \bottomrule
    \end{tabular}
    }
\end{table}

\smallskip

A monitor solves a binary classification problem. Given an input and a network prediction, the task is to either accept or reject this prediction, which is also called a \emph{negative} and a \emph{positive} test, respectively.
Novelty detection corresponds to a \emph{true positive}.
Two types of errors can occur: the input is rejected when it must be accepted (\emph{false positive}, equivalent to a false alarm), or the input is accepted when it must be rejected (\emph{false negative}, equivalent to a miss).
In Tab.~\ref{tbl:test_notation} we outline our notation for these cases later used in the plots.

\begin{table}
    \caption{Notation for possible monitoring outcomes in the plots. We say that a network prediction is correct if it matches the ground truth.
    To save space, we do not show the true negatives; as the plots show percentages, the true negatives are the remaining difference to $100$.}
    \label{tbl:test_notation}
    \centering
    \footnotesize{
    \def\d{@{\hspace*{4mm}}}
    \def\de{@{\hspace*{1.5mm}}}
    \begin{tabular}{\de c \d l \d l \de}
        \toprule
        Pattern & Label & Explanation \\
        \midrule
        \raisebox{-1mm}{\resizebox{5mm}{5mm}{\legendFP}} & false positive & prediction correct $\&$ outside abstraction \\
        \raisebox{-1mm}{\resizebox{5mm}{5mm}{\legendFN}} & false negative & $\neg$ prediction correct $\&$ $\neg$ outside abstraction \\
        \raisebox{-1mm}{\resizebox{5mm}{5mm}{\legendTP}} & true positive & $\neg$ prediction correct $\&$ outside abstraction \\
        \bottomrule
    \end{tabular}
    }
\end{table}

\medskip

Our implementation and trained networks are available online.\footnote{\url{https://github.com/VeriXAI/Outside-the-Box}}

\medskip

In the first experiment, we demonstrate the performance of our monitoring framework in comparison to two related novelty-detection approaches, which we shortly recall next.

\smallskip

The first approach is the \emph{softmax prediction probability} approach following~\cite{HendrycksG17}, which we will call the ``threshold'' approach.
The idea is to interpret the softmax output of the neural network (the normalization of each output value $o_i$ to $o_i/\sum_j o_j$) as a probability.
Given a threshold value $\alpha$, the approach rejects an input whenever the probability assigned to the output class falls below this threshold.
We note that this approach can be restated as a special case in our framework: observing only the output layer, we use a single box for each class $c_i$ with range $[\alpha, 1]$ in dimension $i$ (the dimension corresponding to class $c_i$) and unbounded range $(-\infty, \infty)$ in all other dimensions; since $\alpha$ is fixed, there is no monitor training involved.

\begin{figure}[tb]
	\centering
	\includecutplot[17mm 21mm 22mm 33mm]{keepaspectratio,width=\linewidth}{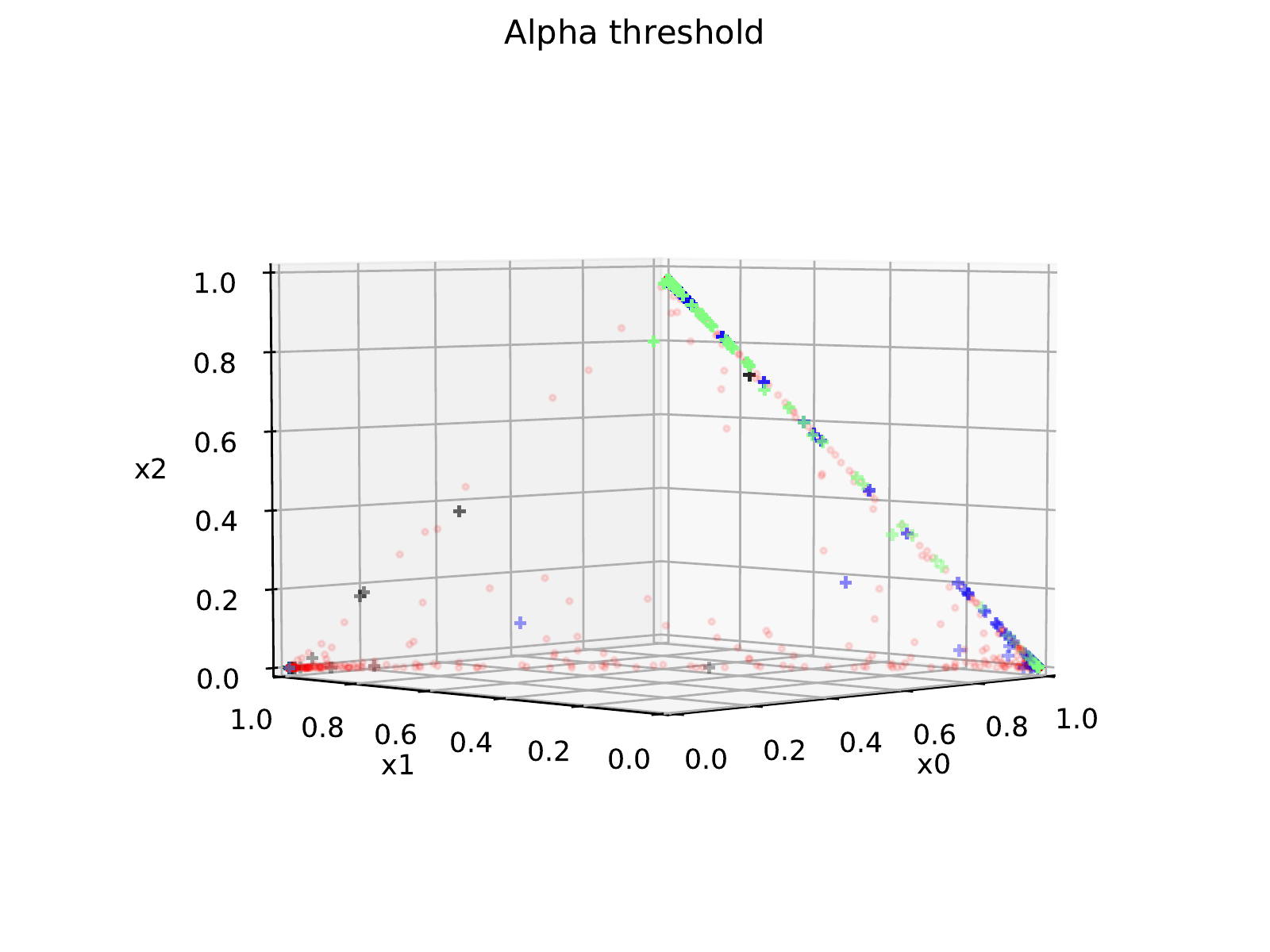}
	\caption{Softmax output for the F\_MNIST benchmark with three known classes (blue ($x_0$), black ($x_1$), green ($x_2$)) and one unknown class (red).
	For each class there are $1{,}000$ inputs.}
	\label{fig:alpha_threshold}
\end{figure}

In Fig.~\ref{fig:alpha_threshold}, we depict the output distribution of the F\_MNIST benchmark for three known and one unknown class.
For three classes the outputs can in principle be distributed on a triangular plane.
However, as can be seen, the network prefers to assign outputs to the corners or at least to the edges of the triangle (``+'' markers in the figure), which correspond to a prediction probability $0$ for two or one of the classes, respectively.
For novel inputs (red dots in the figure) the network tends to do the same.
Hence the threshold approach requires a high value $\alpha$ to reject them (we use 0.9 and 0.99 in the experiments).

We observed that the threshold approach performs poorly in scenarios with few known (or equivalently: many unknown) classes.
Since for $n$ classes the prediction probability can range in the interval $\left[\nfrac{1}{n}\,, 1\right]$, we normalized the threshold values relative to those intervals.
For example, in an experiment with $n = 2$ known classes we map the threshold $0.9$ to $\left[\nfrac{1}{2}\,, 1\right] \cdot 0.9 = 0.95$.
For small values of $n$ the normalized version would thus reject more inputs.
Still, the threshold approach does not perform well in these scenarios.

\smallskip

As the second approach for comparison, we consider what we call the \emph{Boolean abstraction} from~\cite{ChengNY19}.
We apply the abstraction in the last hidden layer, as suggested by the authors.
Given a Boolean abstraction $f$, membership of a vector \vx reduces to satisfiability of a Boolean formula, which is expensive to decide.
Indeed, we were only able to employ the abstraction in the \nnI network's last hidden layer (40 neurons) and ran out of memory (8\,GB) while constructing the formula on the \nnII network's last hidden layer (84 neurons).
Compared to an average training time of $8$ seconds on the $60{,}000$ training inputs and $2$ seconds for monitoring the $10{,}000$ test inputs for our monitor in the MNIST benchmark (with four watched layers), the Boolean abstraction took $48$ seconds for training and $17$ seconds for running, respectively (with a single watched layer).

\begin{figure*}[t]
	\centering
	\begin{subfigure}[t]{.49\linewidth}
		\centering
		\includecutplot[13mm 8mm 21mm 29mm]{keepaspectratio,width=\textwidth,height=\plotheight}{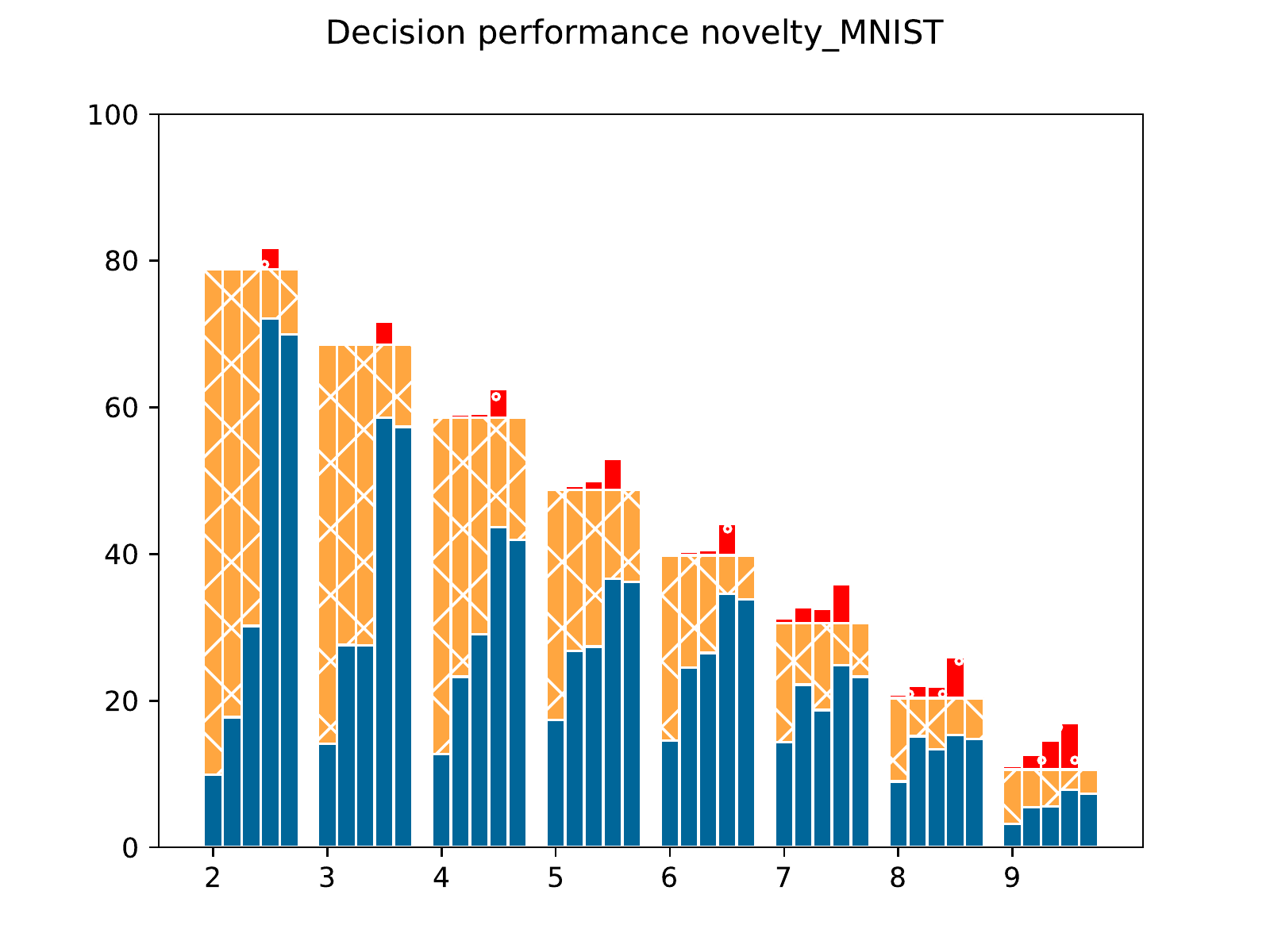}
		\caption{MNIST: last four layers; $\tau = 0.07$; $283$ watched neurons.}
		\label{subfig:comparison_alpha_MNIST}
	\end{subfigure}
	\hfill
	\begin{subfigure}[t]{.49\linewidth}
		\centering
		\includecutplot[13mm 8mm 21mm 29mm]{keepaspectratio,width=\textwidth,height=\plotheight}{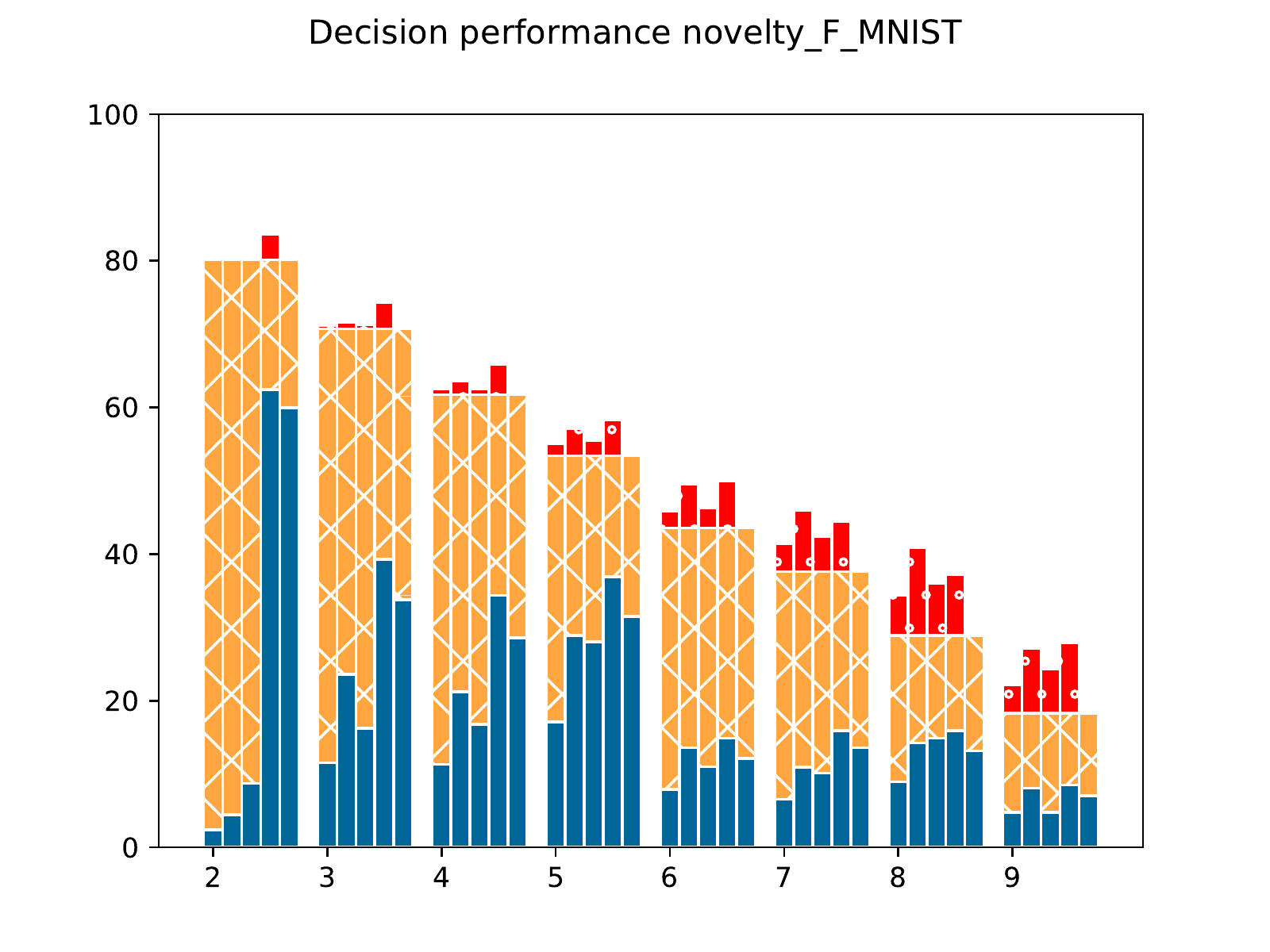}
		\caption{F\_MNIST: last five layers; $\tau = 0.07$; $602$ watched neurons.}
	\end{subfigure}
	\\
	\begin{subfigure}[t]{.49\linewidth}
		\centering
		\includecutplot[13mm 8mm 21mm 29mm]{keepaspectratio,width=\textwidth,height=\plotheight}{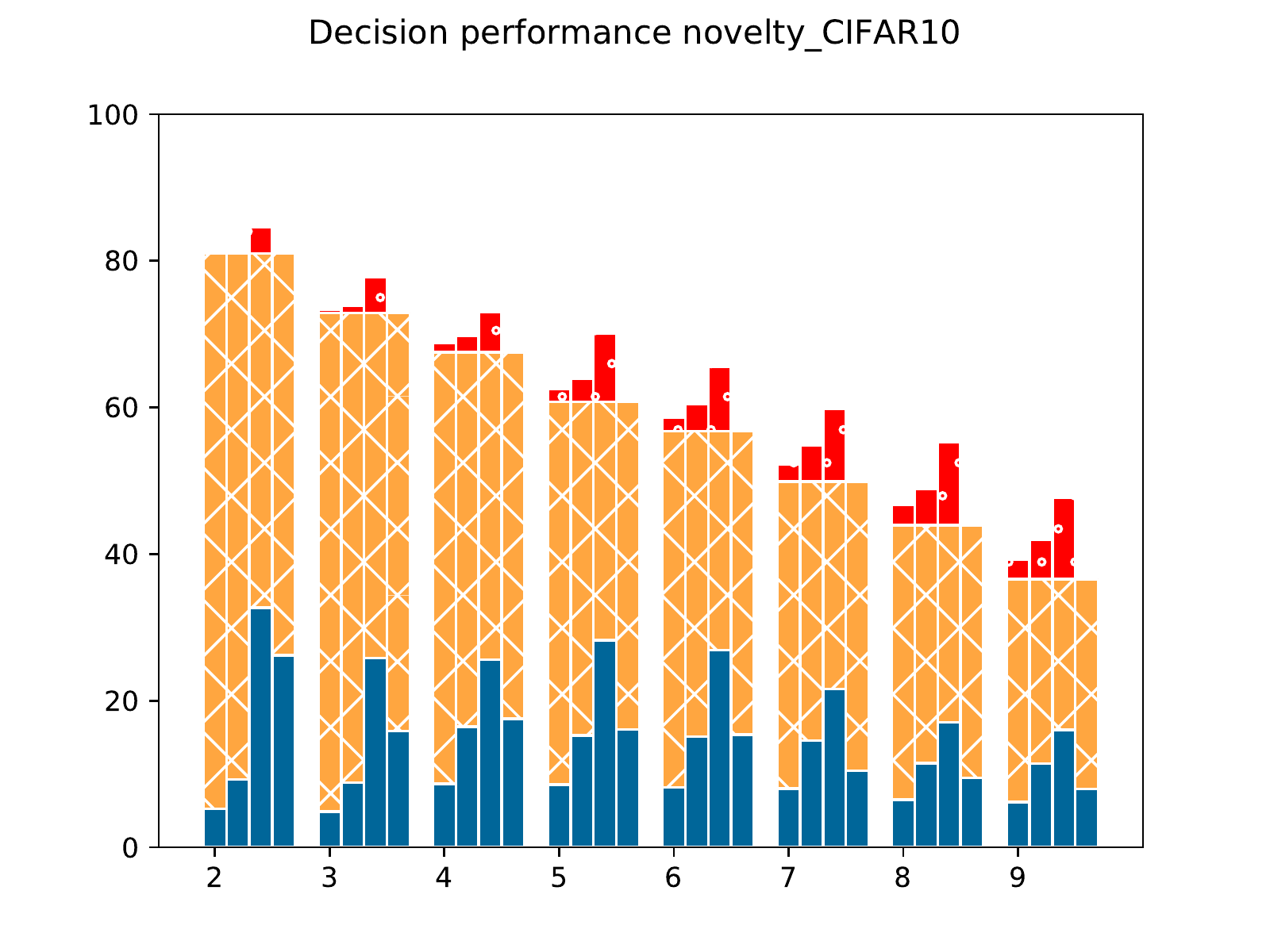}
		\caption{CIFAR-10: last four layers; $\tau = 0.07$; $826$ watched neurons.}
	\end{subfigure}
	\hfill
	\begin{subfigure}[t]{.49\linewidth}
		\centering
		\includecutplot[13mm 8mm 21mm 19mm]{width=\textwidth,height=\plotheight}{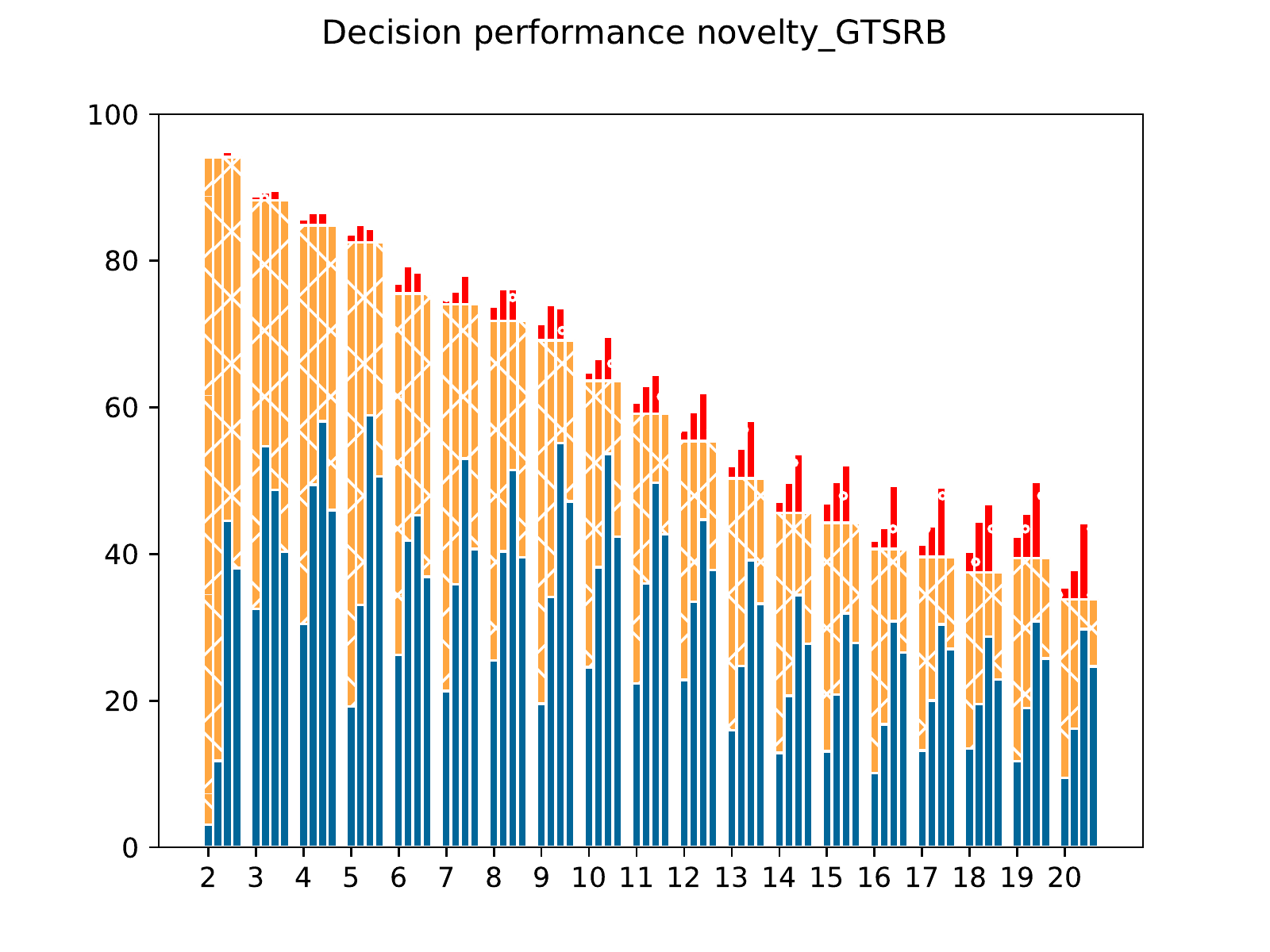}
		\caption{GTSRB: last hidden layer; $\tau = 0.3$; $84$ watched neurons.}
	\end{subfigure}
	\caption{Comparison of novelty detection. Bars from left to right: threshold approach with $\alpha = 0.9$ and $\alpha = 0.99$, Boolean abstraction (MNIST and F\_MNIST only), and box abstraction (excluding and including training on test dataset; parameters given in the captions).
	The x~axis shows the number of known classes and the y~axis shows percentages (see Tab.~\ref{tbl:test_notation}).
	}
	\label{fig:comparison_alpha}
\end{figure*}

\smallskip

In Fig.~\ref{fig:comparison_alpha}, we compare the performance of the outlined approaches.
As can be seen, learning from the test data (last bar) in addition to the training data often barely affects the novelty detection (i.e., the solid blue bars are of almost equal length) but eliminates false warnings (outlined red bars) as expected.
For the first three benchmarks we used a clustering threshold of $\tau = 0.07$.
The GTSRB benchmark (for which we only report results for the instances with up to 20 classes due to lack of space) uses much fewer (less than $1{,}000$) training data per class, which is not enough for our monitor to converge sufficiently.
Hence we used a coarser value $\tau = 0.3$ and also increased the boxes by $10\,\%$ after training.
Notice that for 19 known classes the number of misclassifications (the height of the solid blue and decorated yellow bars combined) produced by the network is higher than the one for the same network trained on 18 known classes, indicating that the former has inferior precision (which we validated on the test dataset).
As the runtime statistics in Table~\ref{tbl:runtime_novelty} show, monitoring takes less than a millisecond per input despite watching multiple layers.

\begin{table}
    \caption{Runtime statistics for different benchmarks.
    We report the accumulated time for clustering (``Cluster'') and for creating the abstraction (``Creation'') on all training inputs, and the time for running the monitor (``Running'') on all test inputs, averaged over all runs presented in Fig.~\ref{fig:comparison_alpha}.
    The results were obtained on a laptop (2.20\,GHz CPU with four cores and 8\,GB RAM) using Linux.}
    \label{tbl:runtime_novelty}
    \centering
    \scriptsize{
    \def\d{@{\hspace*{2mm}}}
    \def\de{@{\hspace*{1mm}}}
    \begin{tabular}{\de r \d r \d r \d r \d r \de}
		\toprule
        & MNIST & F\_MNIST & CIFAR-10 & GTSRB \\
        \midrule
        Cluster & 158.8\,s & 198.1\,s & 168.1\,s & 6.1\,s \\
        Creation & 7.6\,s & 14.3\,s & 15.2\,s & 1.32\,s \\
        Running & 2.1\,s & 4.4\,s & 5.6\,s & 9.2\,s \\
        \bottomrule
    \end{tabular}
    }
\end{table}

We also experimented with combinations of our monitor and the threshold approach. For instance, we used our monitor for rejection and, in case of acceptance, used the result of the threshold approach.
However, we were not able to identify a superior combination, possibly because the threshold approach is a special case of our approach.

\section{FRAMEWORK FLEXIBILITY}

This section is dedicated to demonstrating features of the framework that can be explored by the user to improve the desired metrics.

\subsection{Monitoring multiple layers}

\begin{figure}[t]
	\centering
	\includecutplot[13mm 7mm 21mm 31mm]{keepaspectratio,width=\linewidth,height=\plotheight}{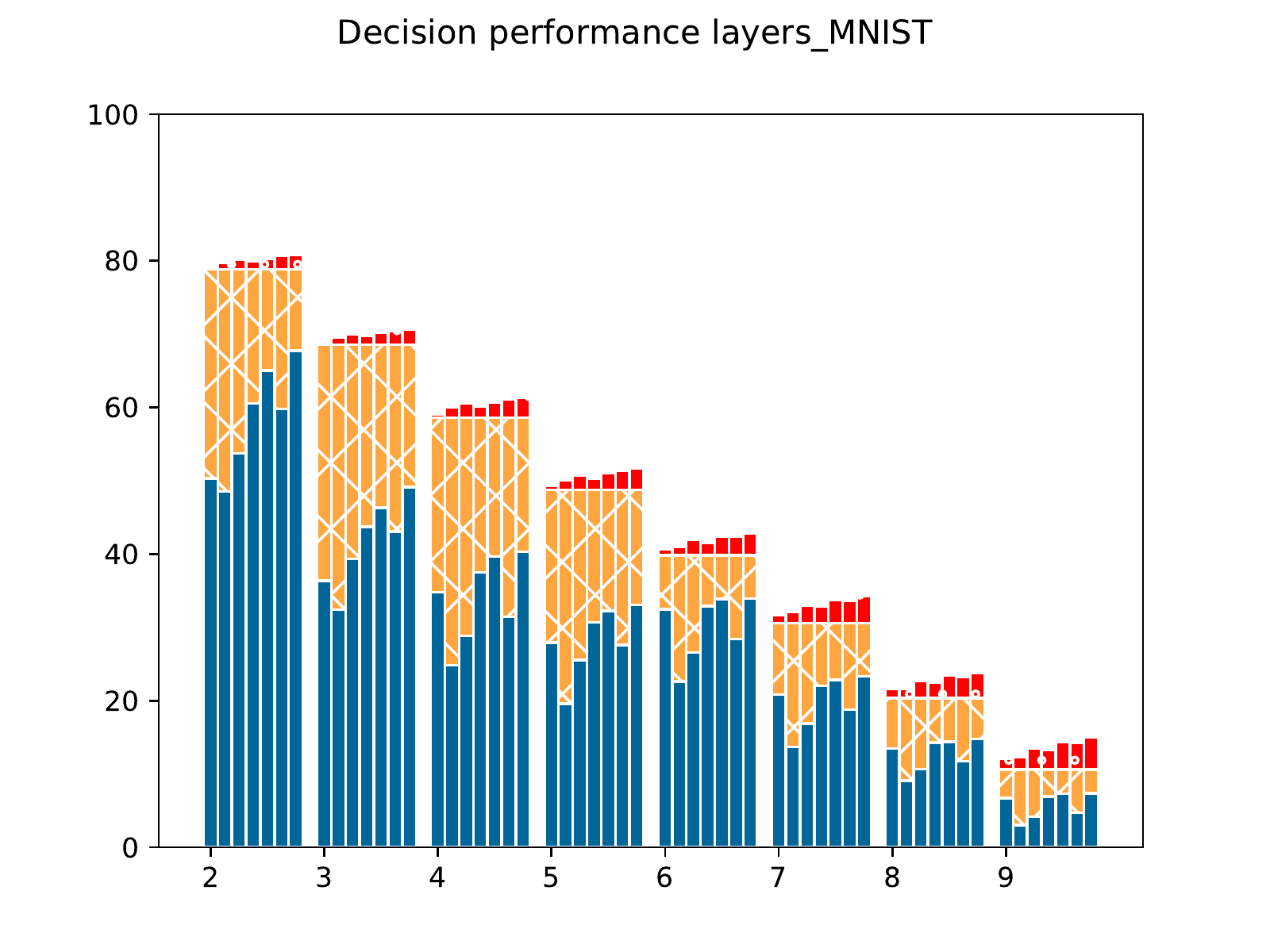}
	\caption{Box abstraction with different observed layers.
	We consider all combinations of the last three layers ($\ell_x$, $\ell_y$, and $\ell_z$) for the MNIST benchmark.
	The order of the bars from left to right is $\ell_z$; $\ell_y$; $\ell_x$; $\ell_y, \ell_z$; $\ell_x, \ell_z$; $\ell_x, \ell_y$; $\ell_x, \ell_y, \ell_z$.}
	\label{fig:layers}
\end{figure}

For evaluating the combination of different layers, we monitored the MNIST benchmark with box abstraction for the last three hidden layers.
We use the monitor policy to accept inputs only if the corresponding outputs at all layers are inside the respective abstraction.
The detection performance is given in Fig.~\ref{fig:layers}.
As expected, the number of warnings increases with the number of watched layers, and we mainly see an increase in correct warnings (true positives).
Interestingly, monitoring the last layer is already sufficiently precise for this benchmark on instances with many known classes (right part of the figure), but on the instances with few known classes the combination with other layers is more powerful and no layer alone can achieve the combined performance (left part of the figure).

\subsection{Enlarging boxes}

\begin{figure}[t]
	\centering
	\raisebox{3mm}{\input{tikz_box_distance}}
	\hfill
	\includecutplot[13mm 8mm 64.3mm 50mm]{keepaspectratio,width=.7\linewidth,height=\plotheight}{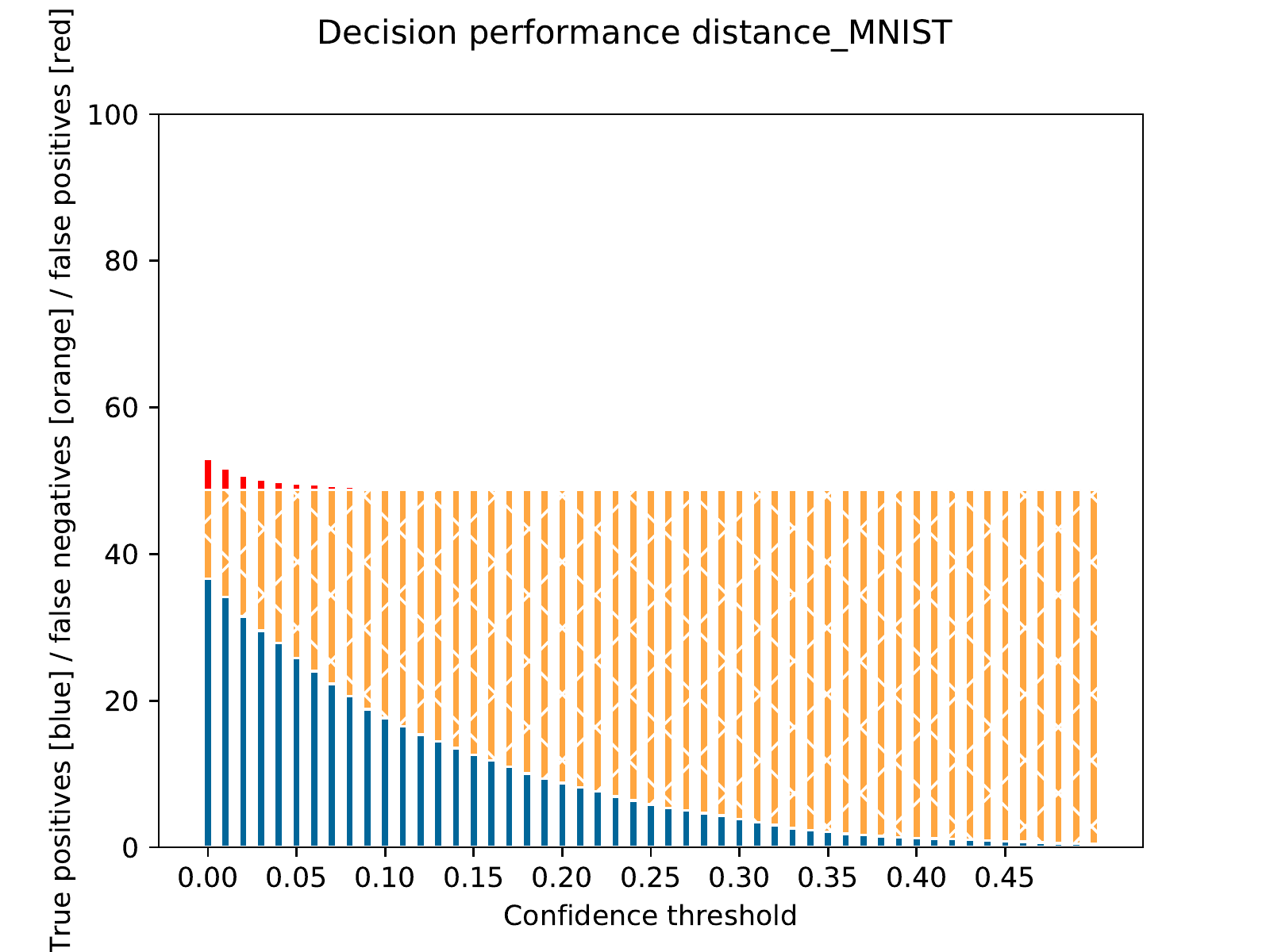}
	\caption{Left: enlarging a box by factor $\gamma$.
	Right: results for enlarged boxes by different factors $\gamma$ (x axis) for the MNIST benchmark with five known and unknown classes each.}
	\label{fig:distance}
\end{figure}

Our monitor can be easily tuned to be more (or less) restrictive by shrinking (or enlarging) the boxes.
This roughly corresponds to de- or increasing the factor $\alpha$ in the threshold approach.
We can simultaneously reason about arbitrary box sizes by computing the Euclidean distance between data points not inside any boxes and the box centers.
We illustrate the idea in Fig.~\ref{fig:distance} on the left.
The line segment connecting the green point with the box center has distance $\gamma d$, where $d$ is the distance between the center and the intersection of the line segment with the border of the box.
Hence we would need to increase the box (i.e., each interval) by at least factor $\gamma$ in order to contain the point (if we use a uniform increase factor).

Having computed the factor $\gamma$ for each point, we plot the results for (uniformly) increased boxes by varying values of $\gamma$ in Fig.~\ref{fig:distance} on the right.
From this plot, based on the preferences, we can choose a policy to achieve, e.g., a fixed false-positive rate.
For instance, to avoid false positives altogether, we should increase the boxes by factor $\gamma = 0.09$ for maximal detection.
Given such a policy, the optimal value $\gamma$ can be found automatically.

\subsection{Comparing abstractions}\label{sec:other_abstractions}

\begin{figure}[t]
	\centering
	\includecutplot[13mm 8mm 21mm 28mm]{keepaspectratio,width=\linewidth,height=\plotheight}{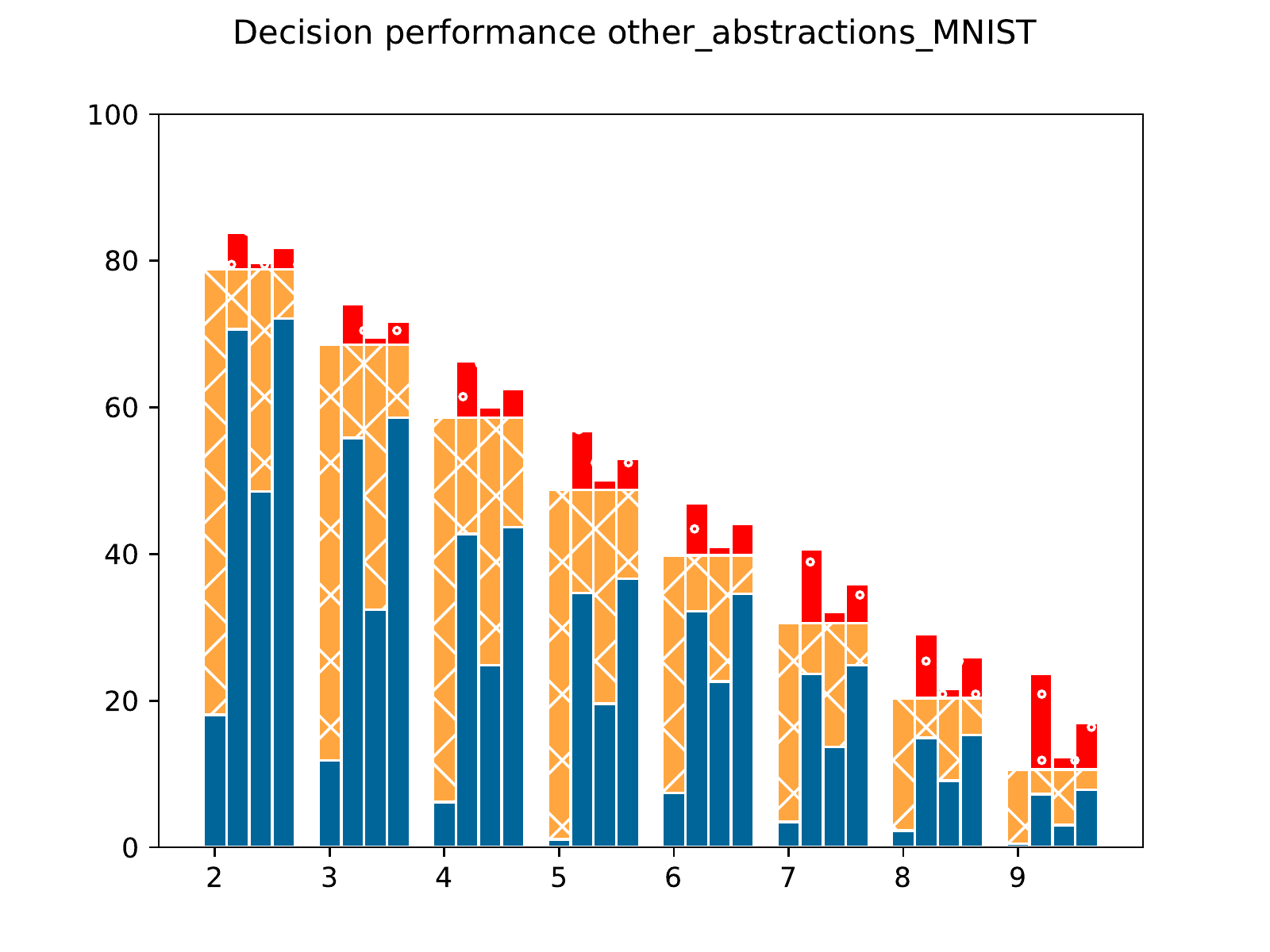}
	\caption{Evaluation of further abstractions on the MNIST benchmark. Bars from left to right: balls, octagons, and boxes in the second-to-last layer, and the bar for boxes from Fig.~\ref{subfig:comparison_alpha_MNIST}.}
	\label{fig:other_abstractions}
\end{figure}

Following the same main ideas as for boxes, our framework allows to consider other types of abstractions.
We report results for two of them: (Euclidean) balls and shapes called ``octagons''~\cite{Mine06}.
Octagons are tighter versions of boxes with additional diagonal constraints between each pair of dimensions.
Octagons can be stored as difference-bound matrices with $\Oof(d^2)$ entries, where $d$ is the number of neurons of the watched layer.
Determining whether a point lies inside a ball is linear in $d$ (as it is for boxes), but for octagons it is quadratic.
Since a ball needs to have the same radius in each dimension and neural networks do not use the same domain for each neuron, balls are too coarse to be effective.
As octagon abstraction is more precise than box abstraction, octagons may detect more novelties but also raise more false warnings.
We compare these abstractions to the box abstraction on the MNIST benchmark in Fig.~\ref{fig:other_abstractions}, where we used the second-to-last layer (first three bars).
In addition, the fourth bar shows the box abstraction from Fig.~\ref{subfig:comparison_alpha_MNIST} in four layers.
As can be seen, boxes in four layers strictly outperform octagons in both higher detection power and fewer false warnings, while being computationally much more efficient ($2$ seconds compared to $121$ seconds).

\section{CONCLUSIONS}

Guaranteeing correctness of systems that rely on neural-network classifiers remains an important open challenge. In safety-critical applications, addressing the problem of novelty detection is crucial. The framework we propose in this paper brings us one step closer to a general methodology for developing reliable machine-learned tools. Inspired by abstraction techniques that have proved effective in the program-analysis domain, we monitor neural networks at runtime. Experimental results on common benchmarks for image classification demonstrate that our framework for constructing box abstractions of neural-network layers is effective in detecting novelties and computationally cheap. As future direction, we plan to apply our approach in a real-world setting such as monitoring neural-network controllers for cyber-physical systems.

\section*{ACKNOWLEDGMENTS}

We thank Christoph Lampert and Nikolaus Mayer for fruitful discussions.
This research was supported in part by
the Austrian Science Fund (FWF) under grants S11402-N23 (RiSE/SHiNE) and Z211-N23 (Wittgenstein Award)
and
the European Union's Horizon 2020 research and innovation programme under the Marie Sk{\l}odowska-Curie grant agreement No.\ 754411.

\clearpage
\bibliography{bibliography}
\bibliographystyle{ecai}

\end{document}

%% file: abstract.tex
Neural networks have demonstrated unmatched performance in a range of classification tasks. Despite numerous efforts of the research community, \emph{novelty detection} remains one of the significant limitations of neural networks. The ability to identify previously unseen inputs as novel is crucial for our understanding of the decisions made by neural networks.
At runtime, inputs not falling into any of the categories learned during training cannot be classified correctly by the neural network.
Existing approaches treat the neural network as a black box and try to detect novel inputs based on the confidence of the output predictions.
However, neural networks are not trained to reduce their confidence for novel inputs, which limits the effectiveness of these approaches.
We propose a framework to monitor a neural network by observing the hidden layers.
We employ a common abstraction from program analysis---boxes---to identify novel behaviors in the monitored layers, i.e., inputs that cause behaviors \emph{outside the box}. For each neuron, the boxes range over the values seen in training.
The framework is efficient and flexible to achieve a desired trade-off between raising false warnings and detecting novel inputs.
We illustrate the performance and the robustness to variability in the unknown classes on popular image-classification benchmarks.

%% file: tikz_framework.tex
\def\vd{6mm}
\def\hd{10mm}
\def\ld{0mm}
\def\id{5mm}
\begin{tikzpicture}[neuron/.style={circle,minimum size=6mm,fill=black},t/.style={->,>=stealth},picture/.style={rectangle,draw},tdash/.style={t,dashed},tdot/.style={t,dashdotted},font=\scriptsize]
	\node[picture] (input) {\includegraphics[keepaspectratio,width=7mm]{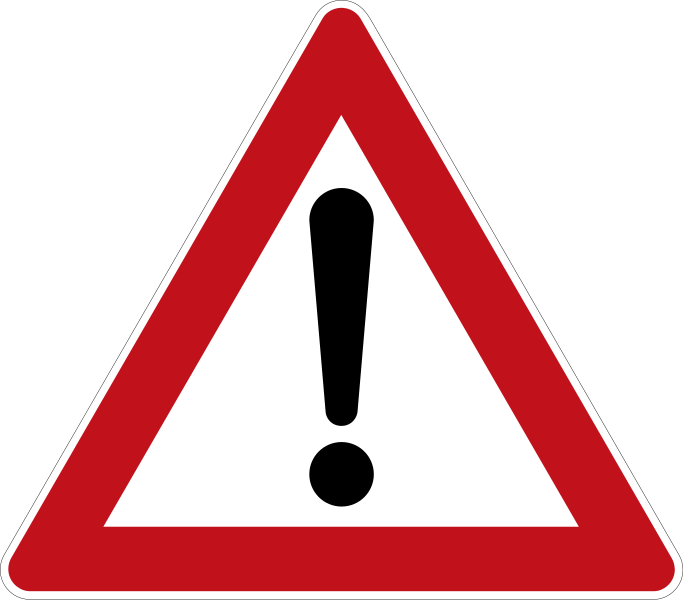}};
	\node[neuron,right=\hd of input,yshift=6mm] (n11) {};
	\node[neuron,below=\vd of n11] (n12) {};
	\node[neuron,right=\hd of n11] (n21) {};
	\node[neuron,below=\vd of n21] (n22) {};
	\node[picture,color=niceorange,fill=niceorange!30,right=\hd of n21] (n31) {\includegraphics[keepaspectratio,width=7mm]{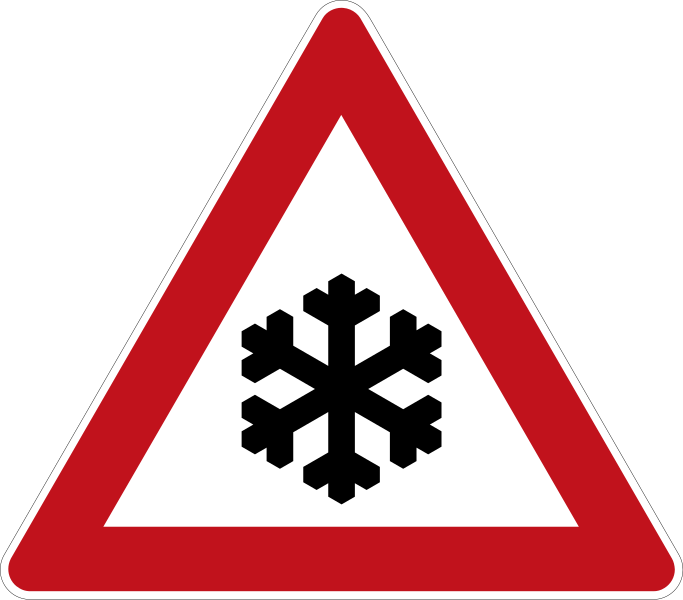}};
	\node[right=0mm of n31] {90\%};
	\node[picture,color=nicegreen,fill=nicegreen!30,right=\hd of n22] (n32) {\includegraphics[keepaspectratio,width=7mm]{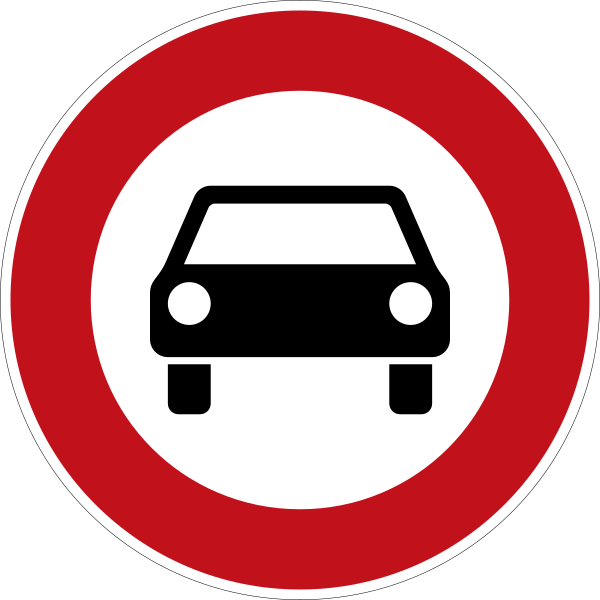}};
	\node[right=0mm of n32] {10\%};
	\coordinate[right=5mm of n11] (nn_dummy);
	\node[above=0mm of n31] (output_label) {\begin{tabular}{@{}c@{}}predicted\\output\end{tabular}};
	\node[at=(nn_dummy|-output_label)] {neural network};
	\node[at=(input|-output_label)] {input};
	\draw[t] (input) to (n11);
	\draw[t] (input) to (n12);
	\draw[t] (n11) to (n21);
	\draw[t] (n11) to (n22);
	\draw[t] (n12) to (n21);
	\draw[t] (n12) to (n22);
	\draw[t] (n21) to (n31);
	\draw[t] (n21) to (n32);
	\draw[t] (n22) to (n31);
	\draw[t] (n22) to (n32);
	\node[opacity=0.5,below=-11mm of n21,xshift=-7mm,xscale=-1] {{\includegraphics[keepaspectratio,height=40mm]{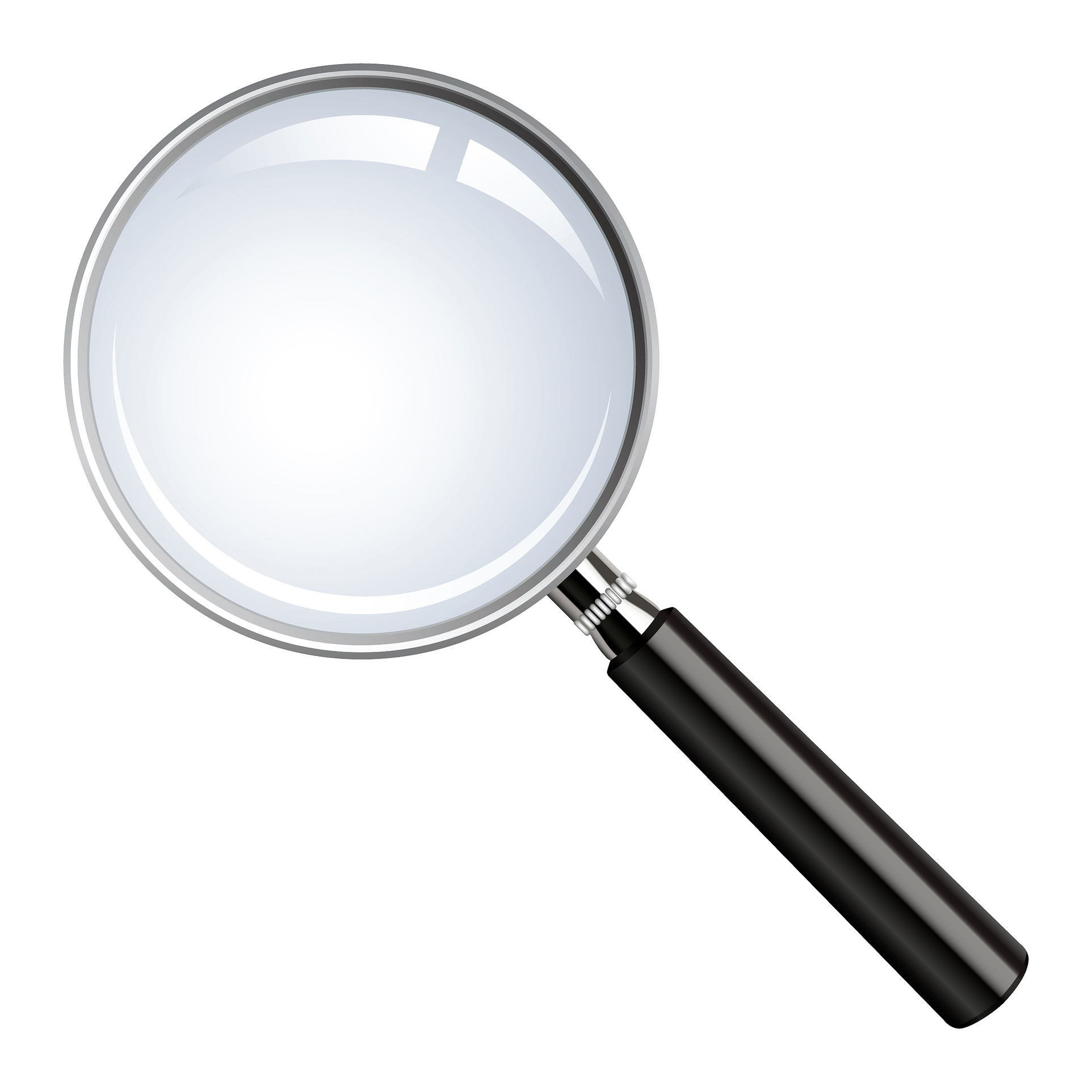}}};
	\node[below=2mm of n12,xshift=-2.5mm,xscale=-1,opacity=0.8] (abstraction) {\includegraphics[keepaspectratio,width=4cm,clip,trim=20.3mm 9.05mm 16mm 10mm]{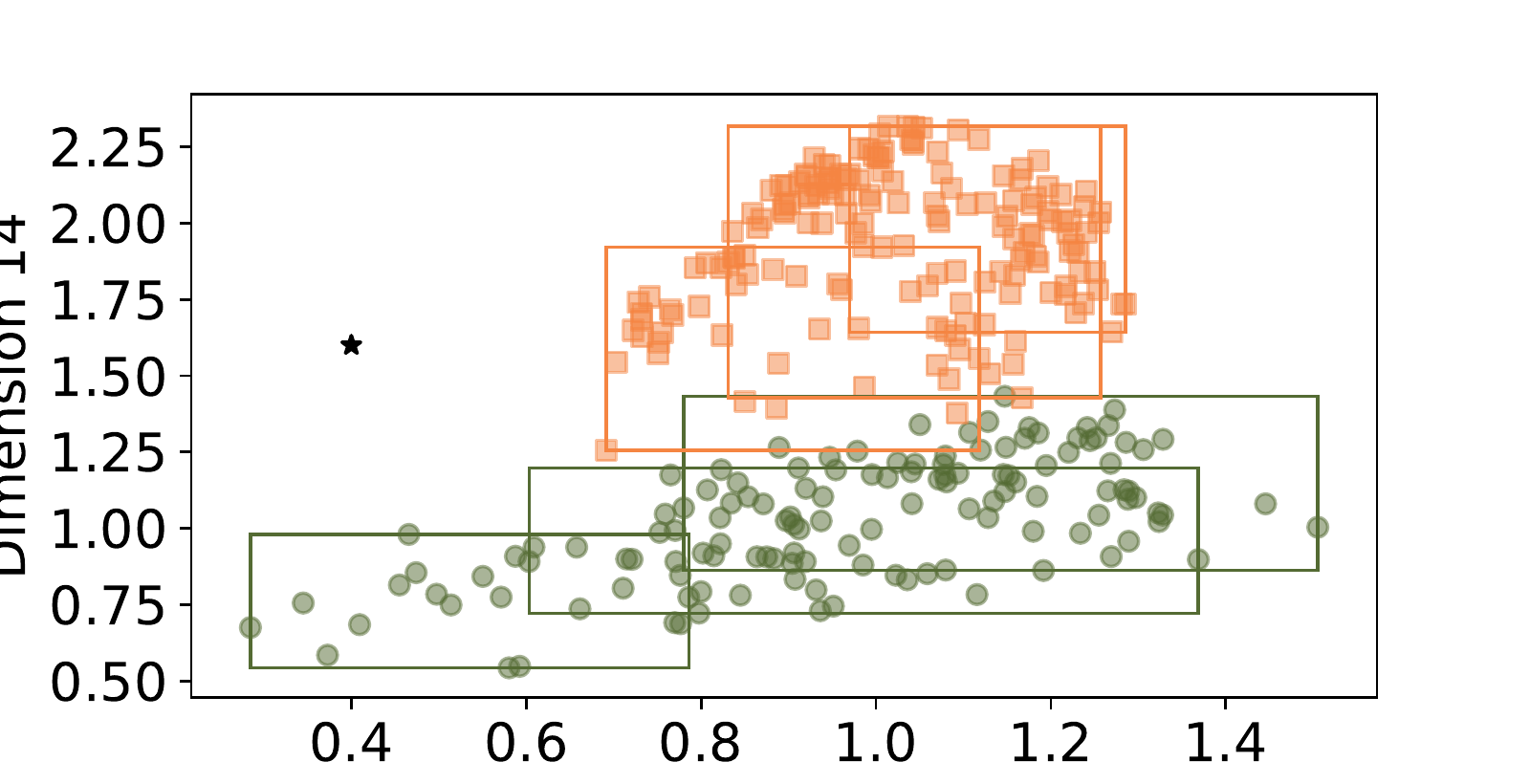}};
	\node[below=-1mm of abstraction] {box abstraction};
	\node at (30mm,-21mm) (outlier) {\phantom{x}};
	\draw[tdot,bend right] (n21) to (outlier);
	\draw[tdot] (n22) to (outlier);
	\node[rectangle,draw,thick,color=black,at=(outlier-|n32)] (unknown) {\textcolor{black}{\begin{tabular}{@{}c@{}}unknown\\input\\detected\end{tabular}}};
	\node[below=-2mm of outlier] (novel_input) {\begin{tabular}{@{}c@{}}novelty\end{tabular}};
	\draw[t,thick] (outlier) to (unknown);
	%
	%
	\node[neuron,color=gray,right=\hd of n11] {};
	\node[neuron,color=gray,below=\vd of n21] {};
\end{tikzpicture}

%% file: tikz_neural_network.tex
\def\vd{6mm}
\def\hd{13mm}
\def\ld{0mm}
\def\id{5mm}
\begin{tikzpicture}[neuron/.style={circle,minimum size=6mm,fill=black},t/.style={->,>=stealth}]
	\node[neuron] (n11) {};
	\node[neuron,below=\vd of n11] (n12) {};
		\node[below=\ld of n12] {$\ell_1$};
	\node[neuron,right=\hd of n11] (n21) {};
	\node[neuron,below=\vd of n21] (n22) {};
		\node[below=\ld of n22] {$\ell_2$};
	\node[neuron,right=\hd of n21] (n31) {};
	\node[neuron,below=\vd of n31] (n32) {};
		\node[below=\ld of n32] {$\ell_3$};
	\draw[t] (n11) to node[above] {$-0.2$} (n21);
	\draw[t] (n11) to node[very near start,below,xshift=-1.5mm,yshift=0.5mm] {$0.3$} (n22);
	\draw[t] (n12) to node[very near end,below,xshift=2mm,yshift=0.5mm] {$0.8$} (n21);
	\draw[t] (n12) to node[below] {$0.6$} (n22);
	\draw[t] (n21) to node[above] {$1.1$} (n31);
	\draw[t] (n21) to node[very near start,below,xshift=-1mm] {$0.1$} (n32);
	\draw[t] (n22) to node[very near end,below,xshift=2mm,yshift=0.5mm] {$0.2$} (n31);
	\draw[t] (n22) to node[below] {$0.8$} (n32);
	\draw [decorate,decoration={brace,amplitude=6pt,raise=4pt,mirror}]
(n11.west) -- (n12.west) node[midway,yshift=0.3mm,xshift=-5mm] {\vx};
	\draw [decorate,decoration={brace,amplitude=6pt,raise=4pt}]
(n31.east) -- (n32.east) node[midway,yshift=-0.3mm,xshift=5mm] {$y$};
\end{tikzpicture}

%% file: tikz_box_distance.tex
\begin{tikzpicture}[r/.style={rectangle,draw,color=nicegreen,thick}]
	\node[r,minimum height=25mm,minimum width=1cm] (box) {};
	\node[r,minimum height=37.5mm,minimum width=15mm,dashed] {};
	\node[circle,fill=black,inner sep=0.6mm,at=(box.center)] (c) {};
	\node[circle,draw=nicegreen,fill=nicegreen!80,inner sep=0.4mm,above=18mm of box.center,xshift=3mm] (p) {};
	\coordinate (m) at (2mm,12.5mm);
	\draw[dashdotted] (c) to node[right] {{$d$}} (m);
	\draw[dotted] (m) to node[right] {{$\gamma d$}} (p);
\end{tikzpicture}